\documentclass[11pt]{article}
\usepackage{fontspec}
\usepackage[a4paper,margin=2cm]{geometry}
\usepackage{graphicx}
\usepackage{booktabs}
\usepackage{amsmath}
\usepackage{amssymb}
\usepackage{hyperref}
\usepackage{xcolor}
\usepackage{enumitem}
\usepackage{multirow}
\usepackage{tabularx}
\usepackage{float}

\title{Systematic Optimization of Real-Time Diffusion Model Inference\\on Apple M3 Ultra}

\author{
Yoichi Ochiai\\
\textit{University of Tsukuba, Faculty of Library, Information and Media Science}\\
\texttt{wizard@slis.tsukuba.ac.jp}
}

\date{}

\begin{document}

\maketitle

\begin{abstract}
While real-time image generation using diffusion models has advanced rapidly on NVIDIA GPUs, systematic optimization research on non-CUDA platforms such as Apple Silicon remains extremely limited.
In this study, we conducted comprehensive optimization experiments across 10 phases targeting the Apple M3 Ultra (60-core GPU, 512\,GB unified memory) with the goal of achieving real-time camera img2img transformation.
We explored a wide range of techniques including CoreML conversion, quantization, Token Merging, Neural Engine utilization, compact model exploration, frame interpolation, kNN search-based synthesis, pix2pix-turbo, optical flow frame skipping, and knowledge distillation, quantitatively evaluating the effectiveness of each approach.
Ultimately, by combining CoreML conversion of the distillation-specialized model SDXS-512 with a 3-thread camera pipeline, we achieved real-time camera img2img transformation at \textbf{22.7 FPS} at 512$\times$512 resolution.
The primary contribution of this work is the systematic demonstration that optimization insights established for CUDA are not necessarily effective on Apple Silicon's unified memory architecture.
We reveal an optimization landscape fundamentally different from that of NVIDIA GPUs---including the absence of speedup from quantization, the ineffectiveness of parallel inference, and the unsuitability of the Neural Engine for large-scale models---and provide practical guidelines for diffusion model inference on Apple Silicon.
\end{abstract}

\section{Introduction}

Diffusion models~\cite{ho2020ddpm,rombach2022ldm} have become the dominant paradigm for text-to-image generation and image transformation.
However, their iterative denoising process is computationally expensive, and real-time inference remains a challenging problem.
In recent years, acceleration techniques have developed rapidly, including single-step distillation approaches such as SD-Turbo~\cite{sauer2023adversarial} and SDXS~\cite{song2024sdxs}, as well as few-step inference methods such as Latent Consistency Models~\cite{luo2023lcm}.
StreamDiffusion~\cite{kodaira2023streamdiffusion} achieved real-time inference exceeding 100 FPS on NVIDIA GPUs through pipeline-level optimization.

However, these acceleration studies almost exclusively assume NVIDIA GPUs and the CUDA ecosystem.
CUDA benefits from decades of accumulated resources including extensive kernel libraries, mature profiling tools, and inference optimization through TensorRT.
In contrast, systematic research on diffusion model optimization for non-CUDA platforms---such as Apple Silicon, Qualcomm Snapdragon, and Intel Arc GPUs---is virtually nonexistent.

The Apple M3 Ultra is a System-on-Chip (SoC) featuring up to 192 GPU cores, 192\,GB of unified memory, and 800\,GB/s memory bandwidth, employing a unique Unified Memory Architecture (UMA) in which the CPU, GPU, and Neural Engine (ANE) share the same memory space.
While this design eliminates the need for CPU--GPU data transfers, it exhibits fundamentally different memory access patterns and computational characteristics from CUDA, meaning that optimization techniques effective on NVIDIA GPUs cannot necessarily be directly applied.

In this study, we conducted systematic optimization experiments across 10 phases on the M3 Ultra (60-core GPU, 512\,GB unified memory configuration) with the goal of achieving real-time camera img2img transformation, starting from StreamDiffusion~\cite{kodaira2023streamdiffusion}.
The contributions of this work are as follows:

\begin{itemize}[leftmargin=*,nosep]
\item A comprehensive benchmark covering more than 10 techniques for diffusion model inference on Apple Silicon
\item Demonstration that CoreML conversion is the only effective UNet acceleration technique, with analysis of the underlying reasons
\item Systematic elucidation of why quantization, Token Merging, parallel inference, and other techniques effective in CUDA environments are ineffective on M3 Ultra
\item Demonstration of the fundamental limitations of replacing diffusion models with kNN search leveraging 512\,GB memory
\item Achievement of 22.7 FPS real-time img2img transformation using the SDXS-512 CoreML pipeline
\item Practical guidelines for diffusion model optimization on unified memory architectures
\end{itemize}

\section{Related Work}

\subsection{Fast Diffusion Models}

Acceleration of Stable Diffusion~\cite{rombach2022ldm} toward real-time inference has been pursued primarily along two directions: model distillation and inference step reduction.

SD-Turbo~\cite{sauer2023adversarial} was the first practical model to achieve single-step inference through Adversarial Diffusion Distillation (ADD), combining adversarial training with score distillation.
SDXS~\cite{song2024sdxs} employs a more advanced distillation approach, achieving high-quality single-step image generation with a lightweight architecture that completely removes the UNet mid-block and reduces the down/up blocks from 4 to 3.
The parameter count is reduced to approximately 38\% (328.2M) of the standard SD-Turbo (865.9M).

Latent Consistency Models (LCM)~\cite{luo2023lcm} enable high-quality generation in 2--4 steps through consistency distillation.
LCM-LoRA~\cite{luo2023lcmlora} makes this distillation applicable as a LoRA adapter, enabling retrofitting to existing models.
Hyper-SD~\cite{ren2024hypersd} combines score distillation with reinforcement learning from human feedback (RLHF), achieving high-quality generation across 1--4 step configurations.
These methods have been primarily evaluated on NVIDIA GPUs, with no reported performance on Apple Silicon.

\subsection{Real-Time Inference Pipelines}

StreamDiffusion~\cite{kodaira2023streamdiffusion} is a framework that optimizes real-time diffusion model inference at the pipeline level.
Its core innovation is the Stream Batch mechanism, which packs different denoising steps of consecutive frames into a single batch to maximize GPU computational parallelism.
For example, in the case of 4-step inference, step 1 of frame $t$, step 2 of frame $t-1$, step 3 of frame $t-2$, and step 4 of frame $t-3$ are processed as a single batch.
Additionally, Residual CFG reuses the Classifier-Free Guidance (CFG) result from the previous frame, halving the number of UNet calls per frame.
These optimizations enable real-time inference exceeding 100 FPS on an NVIDIA RTX 4090.

However, StreamDiffusion's acceleration depends heavily on NVIDIA-specific technologies such as CUDA Streams, TensorRT, and xformers, necessitating alternatives for porting to Apple Silicon.
Furthermore, when using single-step inference (SD-Turbo, SDXS, etc.), the Stream Batch mechanism is inapplicable (there is only one step to batch), making optimization of other pipeline components critical.

\subsection{Machine Learning Inference on Apple Silicon}

Apple has released ml-stable-diffusion~\cite{apple2022mlsd}, a CoreML-based Stable Diffusion inference pipeline.
This framework enables inference on the Neural Engine through a technique called SPLIT\_EINSUM\_V2, which performs split execution of attention operations.
However, SPLIT\_EINSUM\_V2 was designed for the M1/M2 generation Neural Engine and does not fully exploit the high computational performance of M3-generation GPUs.

CoreML is Apple's model inference framework, which converts models from PyTorch or TensorFlow and performs inference using optimized kernels on the Metal GPU backend.
Model conversion via coremltools~\cite{apple2022mlsd} automatically applies static optimization of the computation graph, operator fusion, and memory layout optimization.
In contrast, PyTorch's Metal Performance Shaders (MPS) backend is a general-purpose implementation with limited model-specific optimization.
The performance gap between these two backends has significant implications for diffusion model inference on Apple Silicon.

\subsection{Image Translation Models}

pix2pix-turbo~\cite{parmar2024pix2pix} is an end-to-end img2img translation model based on SD-Turbo that introduces skip connections between the VAE encoder and decoder, directly conveying structural information from the input image to the decoder.
These structural skip connections achieve high structure preservation in transformations from edge-detected images to photorealistic images.
However, the design in which the encoder and decoder share intermediate feature tensors has the side effect of making CoreML conversion of the model as a single subgraph difficult.

\subsection{Retrieval-Based Image Generation}

Retrieval-based image generation leveraging large-scale memory~\cite{blattmann2022retrieval} is an approach that retrieves nearest-neighbor samples from a pre-computed image feature database and uses them as conditioning to assist diffusion model generation.
FAISS~\cite{johnson2019faiss} is a high-speed approximate nearest-neighbor search library that enables sub-millisecond search even on databases of one billion vectors using IVF-PQ indexing.
The M3 Ultra with 512\,GB of unified memory can hold large-scale vector databases in memory that would be infeasible on typical GPUs (24\,GB), suggesting new possibilities for retrieval-based approaches.

\section{Experimental Setup}

The experimental environment used in this study is shown in Table~\ref{tab:env}.
The Apple M3 Ultra is Apple's flagship chip released in 2024, comprising two M3 Max dies interconnected via UltraFusion.
The configuration used in our experiments features a 60-core GPU (16 cores disabled from the maximum 76-core configuration) and 512\,GB of unified memory.
The theoretical FP16 compute performance is approximately 22 TFLOPS, which is roughly 1/15th of the NVIDIA RTX 4090's approximately 330 TFLOPS.
However, the ability of the CPU, GPU, and Neural Engine to share 800\,GB/s of memory bandwidth through unified memory is an advantage unavailable in discrete GPU configurations.

\begin{table}[H]
\centering
\caption{Experimental environment}
\label{tab:env}
\begin{tabular}{ll}
\toprule
Item & Specification \\
\midrule
SoC & Apple M3 Ultra \\
GPU & 60 cores ($\sim$22 TFLOPS FP16) \\
CPU & 32 cores (16P + 16E) \\
Memory & 512\,GB unified memory \\
Memory bandwidth & 800 GB/s \\
Neural Engine & 32 cores \\
OS & macOS Sequoia (Darwin 24.4.0) \\
Python & 3.9 \\
PyTorch & 2.6.0 (MPS backend) \\
CoreML Tools & 9.0 \\
diffusers & 0.36.0 \\
\bottomrule
\end{tabular}
\end{table}

For the software stack, we used PyTorch 2.6.0 with the MPS backend as our baseline and performed model conversion using CoreML Tools 9.0.
Hugging Face diffusers 0.36.0 was used for loading and preprocessing various diffusion models, and OpenCV was used for camera input/output.

\section{Phase 1: Porting to the MPS Backend}

As the first phase, we ported StreamDiffusion, originally implemented exclusively for CUDA, to the Apple Metal Performance Shaders (MPS) backend.
The three primary modifications were:
(1) replacing timing measurements using \texttt{torch.cuda.Event} with \texttt{time.perf\_counter},
(2) changing device specifications from \texttt{cuda} to \texttt{mps}, and
(3) initializing the random number generator on the CPU for reproducibility on MPS before transferring to the GPU.

StreamDiffusion's Stream Batch mechanism is not applicable in our experiments using single-step inference (SD-Turbo).
This is because Stream Batch is designed to parallelize multiple steps, and with single-step inference there are no steps to batch.
Consequently, the optimization focus in this study is directed toward accelerating the entire single-frame inference pipeline.

Baseline performance after porting is shown in Table~\ref{tab:baseline}.
Inference of SD-Turbo (865.9M parameters) on MPS required 95.8\,ms/frame (10.4 FPS) at 512$\times$512 resolution.
Even at a reduced resolution of 256$\times$256, it only achieved 79.5\,ms (12.6 FPS), indicating that the speed improvement was limited relative to the reduction in computation, suggesting that simple resolution reduction alone is insufficient.

\begin{table}[H]
\centering
\caption{Baseline performance after MPS porting (SD-Turbo, 1-step)}
\label{tab:baseline}
\begin{tabular}{ccc}
\toprule
Resolution & ms/frame & FPS \\
\midrule
256$\times$256 & 79.5 & 12.6 \\
384$\times$384 & 87.8 & 11.4 \\
512$\times$512 & 95.8 & 10.4 \\
\bottomrule
\end{tabular}
\end{table}

\section{Phase 2: Comprehensive Evaluation of Acceleration Techniques}

In this phase, we systematically evaluated multiple optimization techniques whose effectiveness has been reported on NVIDIA GPUs on the Apple M3 Ultra.
To state the conclusion upfront, only CoreML conversion proved effective; all other techniques were either ineffective or counterproductive.

\subsection{CoreML Conversion}

We converted the PyTorch model to CoreML format (.mlpackage) and performed inference on the Metal GPU backend.
Conversion was performed using trace-based conversion via \texttt{ct.convert} with \texttt{compute\_units=CPU\_AND\_GPU} specified.
During conversion, static analysis of the computation graph for operator fusion, direct mapping to Metal kernels, and memory layout optimization are automatically applied.

As shown in Table~\ref{tab:coreml}, CoreML conversion reduced UNet inference time from 87.6\,ms to 53.4\,ms, a 39\% reduction.
This improvement is attributed to the effect of CoreML's optimized Metal kernels that account for model structure, compared to the general-purpose Metal Shader implementation of PyTorch's MPS backend.
Throughout this study, \textbf{CoreML conversion was found to be the only effective acceleration technique for UNet inference on Apple Silicon}.

\begin{table}[H]
\centering
\caption{UNet inference acceleration via CoreML conversion}
\label{tab:coreml}
\begin{tabular}{lcc}
\toprule
Backend & UNet inference & FPS \\
\midrule
MPS (PyTorch) & 87.6ms & 11.4 \\
\textbf{CoreML (CPU\_AND\_GPU)} & \textbf{53.4ms} & \textbf{18.7} \\
\bottomrule
\end{tabular}
\end{table}

\subsection{Quantization}

We comprehensively evaluated CoreML's post-training quantization options (INT8 linear, 6-bit palettization, 4-bit palettization, and 2-bit palettization).
On NVIDIA GPUs, inference acceleration through memory bandwidth reduction using techniques such as TensorRT INT8 quantization and AWQ/GPTQ has been widely reported.

However, as shown in Table~\ref{tab:quant}, no change in inference speed was observed at any quantization level on the M3 Ultra.
The fact that even 2-bit palettization (theoretically 1/16th memory bandwidth) produced no acceleration strongly suggests that M3 Ultra GPU inference for UNet is compute-bound rather than memory-bandwidth-bound.
This is likely because the unified memory's 800\,GB/s bandwidth provides sufficient headroom for transferring the weights of an 865.9M-parameter model.

\begin{table}[H]
\centering
\caption{Effect of CoreML quantization (UNet, 512$\times$512)}
\label{tab:quant}
\begin{tabular}{lcc}
\toprule
Quantization scheme & UNet inference & Speedup \\
\midrule
FP16 (baseline) & 53.2ms & --- \\
INT8 linear & $\sim$53ms & 0\% \\
6-bit palettize & $\sim$53ms & 0\% \\
4-bit palettize & $\sim$53ms & 0\% \\
2-bit palettize & $\sim$53ms & 0\% \\
\bottomrule
\end{tabular}
\end{table}

\subsection{Token Merging}

Token Merging (ToMe)~\cite{bolya2023tome} is a technique that merges tokens based on similarity between Key tokens in self-attention, reducing the computational cost of the attention operation.
Its effectiveness has been reported on NVIDIA GPUs when the attention operation is the bottleneck.

However, on MPS, the overhead of token similarity computation and merging operations exceeded the reduction in attention computation, resulting in an \textbf{approximately 10\% slowdown}.
This suggests that the attention implementation on MPS has different bottleneck characteristics compared to xformers and Flash Attention on NVIDIA GPUs.
That is, on MPS, the attention operation itself is not the primary bottleneck, and the additional cost of token manipulation is relatively large.

\subsection{CoreML Parallel Inference}

To maximize utilization of the M3 Ultra's 60-core GPU, we attempted parallel inference using multiple CoreML model instances.
On NVIDIA GPUs, parallel execution of multiple kernels via CUDA Streams is possible and widely used for batch inference and pipeline parallelization.

However, with CoreML, no throughput improvement was observed with 1--4 instances of parallel execution.
CoreML serializes inference requests on the Metal Command Queue, and simultaneous execution of multiple models does not translate to GPU-level parallelism.
This represents a fundamental difference from NVIDIA GPUs, where low-level parallel control via CUDA Streams is possible, indicating that model-level parallelization strategies cannot be applied on Apple Silicon.

\subsection{Neural Engine}

To leverage the 32-core Neural Engine (ANE) of the M3 Ultra, we evaluated different CoreML compute unit configurations.
The ANE is a dedicated processor specialized for power-efficient matrix operations, known to demonstrate superior power efficiency over GPUs for small-scale model inference.

As shown in Table~\ref{tab:ane}, the ANE alone (CPU\_AND\_NE) required 329.4\,ms for UNet inference---6.2$\times$ slower than the GPU alone.
Even with mixed GPU+ANE execution (ALL), the result was 63.6\,ms, 19\% slower than GPU alone (53.2\,ms).
A UNet with 865.9M parameters far exceeds the ANE's processing capacity, and the overhead of data partitioning, transfer, and recombination likely outweighs any benefit from parallelized computation.

\begin{table}[H]
\centering
\caption{UNet inference speed by compute unit}
\label{tab:ane}
\begin{tabular}{lcc}
\toprule
Compute unit & UNet inference & Relative to GPU \\
\midrule
\textbf{CPU\_AND\_GPU} & \textbf{53.2ms} & \textbf{1.0$\times$} \\
ALL (GPU+ANE) & 63.6ms & 1.2$\times$ \\
CPU\_AND\_NE (ANE) & 329.4ms & 6.2$\times$ \\
\bottomrule
\end{tabular}
\end{table}

\subsection{Other Techniques}

\texttt{torch.compile} crashed with runtime errors on the MPS backend and could not be evaluated.
This reflects the fact that PyTorch's compiler stack is optimized for CUDA/CPU and has insufficient support for the MPS backend.
Attention Slicing caused approximately 40\% slowdown, likely because the memory management overhead between slices on MPS outweighs the VRAM savings.
Switching to FP32 precision produced no change in speed, and construction of an asynchronous pipeline yielded no practical improvement.

\subsection{Phase 2 Summary}

Table~\ref{tab:phase2} summarizes the effects of all techniques evaluated in Phase 2.
Only CoreML conversion achieved a 64\% speedup; all other techniques were either ineffective or counterproductive.
This result systematically demonstrates that many optimization techniques established in CUDA environments are inapplicable on Apple Silicon.

\begin{table}[H]
\centering
\caption{Comprehensive evaluation of acceleration techniques}
\label{tab:phase2}
\begin{tabular}{lcc}
\toprule
Technique & Effect & Notes \\
\midrule
\textbf{CoreML conversion} & \textbf{+64\%} & Only effective method \\
Quantization (INT8--2bit) & 0\% & Compute-bound \\
Token Merging & $-$10\% & MPS overhead \\
Parallel inference & 0\% & Metal serialization \\
Neural Engine & $-$19\%\textasciitilde$-$520\% & Unsuitable for large UNet \\
torch.compile & --- & Unsupported on MPS \\
Attention Slicing & $-$40\% & Inefficient on MPS \\
\bottomrule
\end{tabular}
\end{table}

\section{Phase 3: Compact Models and Resolution Optimization}

Since Phase 2 revealed that CoreML conversion is the only effective UNet acceleration technique, we next explored acceleration through reducing model parameter counts.
We converted and evaluated Small-SD (579.4M) and Tiny-SD (323.4M), made available through the Knowledge Distillation Stable Diffusion project, via CoreML conversion.

As shown in Table~\ref{tab:models}, parameter reduction directly translated to inference speed, with Tiny-SD achieving 1.7$\times$ the speed of SD-Turbo.
At a resolution reduced to 320$\times$320, UNet inference alone reached 16.4\,ms (61 FPS), but the quality of fine details in generated images degraded significantly.
For SD-Turbo-family models, deviation from the training resolution (512$\times$512) directly results in quality degradation, leading us to conclude that resolution reduction is not a viable strategy.

\begin{table}[H]
\centering
\caption{Relationship between model size and inference speed (CoreML, 512$\times$512)}
\label{tab:models}
\begin{tabular}{lccc}
\toprule
Model & Parameters & UNet inference & UNet FPS \\
\midrule
SD-Turbo & 865.9M & 53.2ms & 18.8 \\
Small-SD & 579.4M & 36.6ms & 27.3 \\
Tiny-SD & 323.4M & 31.3ms & 31.9 \\
\bottomrule
\end{tabular}
\end{table}

It should be noted that Small-SD and Tiny-SD were unable to maintain the same quality as SD-Turbo through the distillation process, and their subjective image quality in camera img2img transformation was inferior to SD-Turbo.
While parameter reduction contributes to inference speed, maintaining distillation quality remains a challenge.
This insight leads to the superiority of SDXS-512 (designed specifically for distillation with maintained quality) in the subsequent Phase 6.

\section{Phase 4: Camera img2img Pipeline Construction}

\subsection{Bottleneck Analysis}

To achieve real-time camera img2img transformation, we profiled the entire pipeline.
The pipeline consists of camera capture $\to$ preprocessing (resize, normalization) $\to$ VAE encode $\to$ noise addition $\to$ UNet inference $\to$ VAE decode $\to$ postprocessing (display).

As shown in Table~\ref{tab:pipeline}, the UNet constitutes a clear bottleneck, accounting for 68\% of the total inference time.
VAE encoding and decoding used CoreML-converted TAESD (Tiny Autoencoder for Stable Diffusion), achieving fast execution at 6.5\,ms each.
Preprocessing (camera frame resize and Canny edge detection) required 7.9\,ms, and postprocessing (tensor-to-BGR conversion and display) required 2.4\,ms.

\begin{table}[H]
\centering
\caption{Camera pipeline time breakdown (SD-Turbo CoreML)}
\label{tab:pipeline}
\begin{tabular}{lcc}
\toprule
Stage & Time & Proportion \\
\midrule
Preprocessing (resize, canny) & 7.9ms & 10\% \\
VAE Encode (CoreML TAESD) & 6.5ms & 8\% \\
\textbf{UNet (CoreML)} & \textbf{53.2ms} & \textbf{68\%} \\
VAE Decode (CoreML TAESD) & 6.5ms & 8\% \\
Postprocessing (display) & 2.4ms & 3\% \\
\midrule
Total & 77.7ms & 12.9 FPS \\
\bottomrule
\end{tabular}
\end{table}

\subsection{Frame Interpolation and Smoothing}

To reduce the execution frequency of the UNet, we attempted intermediate frame generation through linear interpolation between UNet frames and the previous frame.
However, the quality difference between UNet-generated frames and interpolated frames was large, causing severe visual oscillation (flicker) that prevented achieving practical quality.
EMA (Exponential Moving Average) smoothing reduced oscillation but introduced ghosting artifacts for fast-moving subjects.

\subsection{3-Thread Camera Architecture}

Ultimately, we adopted an architecture that executes camera acquisition, inference, and display in three independent threads.
The inference thread continuously processes the latest camera frame, while the display thread shows the latest inference result.
Temporal coherence between frames was ensured through a combination of fixed noise seeds, feedback of the previous latent output ($\alpha=0.3$), and EMA smoothing.
This configuration achieved flicker-free real-time camera img2img transformation at 13.8 FPS with SD-Turbo CoreML.

\section{Phase 5: Exploring Model Quality Improvements}

By Phase 4, SD-Turbo CoreML had achieved 13.8 FPS, but we explored several approaches to further improve generation quality.

Applying Apple's official SPLIT\_EINSUM\_V2 conversion resulted in a 167\% slowdown (5.2 FPS) due to wasted computation caused by the design enforcing batch=2.
Applying the Hyper-SD 1.5 LoRA adapter degraded the stability of single-step inference with no quality improvement observed.
SDXL-class models (2.6B parameters) required over 200\,ms for UNet inference even after CoreML conversion, making real-time performance infeasible.

These experiments led to the insight that quality improvement requires not adapter additions to existing models or the use of larger models, but rather models designed from the outset to balance inference efficiency with quality.
This perspective guided the selection of SDXS-512 in Phase 6.

\section{Phase 6: Acceleration with SDXS-512}

\subsection{SDXS-512 Design}

SDXS-512~\cite{song2024sdxs} is a model distilled specifically for single-step inference, with the following design features:
(1) complete removal of the UNet mid-block to reduce computation,
(2) reduction of down/up blocks from the standard 4 stages to 3, and
(3) training during distillation to maximize image quality in a single step.
As a result, the parameter count is reduced to 328.2M (38\% of SD-Turbo).

\subsection{Performance Evaluation}

As shown in Table~\ref{tab:sdxs}, CoreML conversion of SDXS-512 achieved 24.4\,ms for UNet inference (2.2$\times$ faster than SD-Turbo) and a camera FPS of 22.7 (1.6$\times$ that of SD-Turbo).
Compared with Tiny-SD (323.4M, similar parameter count) evaluated in Phase 3, SDXS-512 demonstrated significantly superior image quality at comparable speed.
This is attributable to SDXS-512's design, which achieves both architectural reduction and maintained distillation quality.

\begin{table}[H]
\centering
\caption{SDXS-512 vs SD-Turbo (CoreML, 512$\times$512, camera)}
\label{tab:sdxs}
\begin{tabular}{lcccc}
\toprule
Model & Parameters & UNet & UNet FPS & Camera FPS \\
\midrule
SD-Turbo & 865.9M & 53.6ms & 18.7 & 13.8 \\
\textbf{SDXS-512} & \textbf{328.2M} & \textbf{24.4ms} & \textbf{41.0} & \textbf{22.7} \\
\bottomrule
\end{tabular}
\end{table}

The 22.7 FPS in the camera pipeline corresponds to a total processing time of 44.1\,ms per frame.
The breakdown is UNet inference at 24.4\,ms, VAE encode/decode at approximately 5\,ms each, and pre/postprocessing at approximately 10\,ms.
Thanks to the 3-thread architecture, which parallelizes inference with camera acquisition and display, the perceived video was even smoother.

\section{Phase 7: kNN Search-Based Image Synthesis}

\subsection{Research Hypothesis and Motivation}

The M3 Ultra's 512\,GB unified memory is more than 20$\times$ the capacity of typical GPUs (12--24\,GB VRAM).
To fully exploit this large memory capacity, we tested the hypothesis of replacing the UNet's ``generation through computation'' with ``retrieval from memory.''
Specifically, we attempted to pre-generate and store a large number of (input, output) pairs and, for new inputs, search for and interpolate the most similar outputs from the database, thereby completely bypassing UNet inference.

\subsection{FAISS Search Speed Evaluation}

We first evaluated kNN search speed against large-scale vector databases.
As shown in Table~\ref{tab:faiss}, for 768-dimensional CLIP embedding vectors, approximate search using IVF-PQ indexing achieved extremely fast search times of 0.50\,ms even at 100 million vectors.
Even exact search (Flat Index) achieved 49.6\,ms at 1 million vectors, and the advantage of 512\,GB memory was clearly demonstrated in terms of search speed compared to the infeasibility of exact search on 10 million vectors with a 24\,GB GPU due to memory limitations.

\begin{table}[H]
\centering
\caption{FAISS kNN search speed (768-dim CLIP embedding)}
\label{tab:faiss}
\begin{tabular}{rcc}
\toprule
Database size & Flat (exact) & IVF-PQ (approximate) \\
\midrule
1,000 & 0.04ms & --- \\
10,000 & 0.45ms & --- \\
100,000 & 4.75ms & --- \\
1,000,000 & 49.6ms & 0.48ms \\
10,000,000 & 490ms & 0.47ms \\
100,000,000 & --- & 0.50ms \\
\bottomrule
\end{tabular}
\end{table}

\subsection{Image Synthesis Trials}

We attempted image synthesis using three approaches.
(a) \textbf{CLIP-kNN}: Search by CLIP embedding of the input image and compute a weighted average of the top-$k$ output images in latent space. While fast at 26.1\,ms (38.3 FPS), blurring from weighted averaging and temporal instability (oscillation) of search results were problematic.
(b) \textbf{VAE Latent kNN}: Direct search and interpolation in VAE latent space. This was the fastest at 22.1\,ms (45.3 FPS), but search accuracy was poor due to the lack of semantic structure in the latent space, resulting in significantly degraded output quality.
(c) \textbf{Hybrid}: Use kNN search results as the initial latent for UNet, refined with one step of UNet inference. Quality improved, but the addition of CLIP encoding overhead resulted in slowdown to 93.6\,ms (10.7 FPS).

\begin{table}[H]
\centering
\caption{Results of kNN search-based image synthesis}
\label{tab:knn}
\begin{tabular}{lccc}
\toprule
Method & Speed & FPS & Quality \\
\midrule
CLIP-kNN & 26.1ms & 38.3 & $\times$ \\
VAE Latent kNN & 22.1ms & 45.3 & $\times$ \\
Hybrid & 93.6ms & 10.7 & $\triangle$ \\
\bottomrule
\end{tabular}
\end{table}

\subsection{Fundamental Limitations}

The failure to achieve practical quality across all approaches stems from the fundamental difference between kNN search and diffusion models.
The UNet in a diffusion model is a nonlinear function approximator with hundreds of millions of parameters that generates continuous and smooth outputs for given inputs.
In contrast, kNN search approximates from a finite set of discrete samples: (1) weighted averaging in latent space destroys the structural coherence of individual samples, (2) it is fundamentally impossible to pre-compute coverage of the infinite combinations of camera inputs (lighting, composition, subject), and (3) the lack of guaranteed temporal consistency in search results causes oscillation, where search results change discontinuously between frames.

\section{Phase 8: pix2pix-turbo}

pix2pix-turbo~\cite{parmar2024pix2pix} is an SD-Turbo-based model specialized for edge-to-image tasks, achieving high preservation of input structure through skip connections between the VAE encoder and decoder.
Since the UNet component is equivalent to the standard SD-Turbo (865.9M), CoreML conversion yielded an inference speed of 52.9\,ms.

However, the skip-connected VAE, in which the encoder and decoder share intermediate feature tensors, is incompatible with CoreML's static computation graph conversion.
Consequently, VAE encoding and decoding remained as PyTorch inference on the MPS backend, requiring approximately 160\,ms total (encode $\sim$80\,ms + decode $\sim$80\,ms).
As a result, the VAE became a bottleneck 3$\times$ larger than the UNet (53\,ms), limiting camera FPS to 4.0 (Table~\ref{tab:pix2pix}).

\begin{table}[H]
\centering
\caption{pix2pix-turbo performance (512$\times$512)}
\label{tab:pix2pix}
\begin{tabular}{lcc}
\toprule
Configuration & Pipeline time & Camera FPS \\
\midrule
Full MPS (FP16) & 339ms & 3.0 \\
CoreML UNet + MPS VAE & $\sim$250ms & 4.0 \\
\midrule
\multicolumn{3}{l}{\textit{Breakdown: UNet 53ms + VAE 160ms + pre/postprocessing 37ms}} \\
\bottomrule
\end{tabular}
\end{table}

Replacing the VAE with TAESD is not possible because TAESD lacks the skip connection structure, and we concluded that pix2pix-turbo is not suitable for real-time operation on Apple Silicon in its current form.
This result illustrates the important lesson that model architecture design directly affects the feasibility of hardware optimization.

\section{Phase 9: Optical Flow Frame Skipping}

Rather than executing the UNet on every frame, we evaluated a technique that runs the UNet only once every $N$ frames, with the intermediate $(N-1)$ frames complemented by warping the previous frame using optical flow.
Theoretically, for $N=3$, the average of a UNet frame (51.7\,ms) and warp frames (6.6\,ms $\times$ 2) yields $(51.7 + 6.6 \times 2) / 3 = 21.6$\,ms/frame $\approx$ 46 FPS.

The Farneback method was used for optical flow, and to reduce computational cost, the input resolution was downscaled from 512$\times$512 to 256$\times$256 for flow field computation, then upscaled to 512$\times$512 for warping.
This half-resolution flow computation reduced warping time from 22.3\,ms to 6.6\,ms (Table~\ref{tab:flow}).

\begin{table}[H]
\centering
\caption{Optical flow frame skipping performance ($N=3$)}
\label{tab:flow}
\begin{tabular}{lcc}
\toprule
Frame type & Processing time & Notes \\
\midrule
UNet frame & 51.7ms & Full pipeline \\
Warp frame (512$\times$512 flow) & 22.3ms & Farneback \\
Warp frame (256$\times$256 flow) & 6.6ms & Half resolution \\
\midrule
Overall ($N=3$, half-res flow) & --- & \textbf{17.4 FPS} \\
\bottomrule
\end{tabular}
\end{table}

However, the measured result was only 17.4 FPS, merely 38\% of the theoretical value (46 FPS).
This discrepancy is attributable to synchronization overhead in the single-threaded implementation.
Furthermore, since warp frames are mere image deformations rather than AI-generated outputs, pronounced jelly-like distortions occurred in regions of large motion, and the subjective image quality was significantly inferior to the SDXS baseline (22.7 FPS).
As this approach was inferior to the SDXS baseline in both speed and quality, it was not adopted.

\section{Phase 10: Knowledge Distillation for Direct Transformation}

We evaluated a technique that uses the entire SDXS pipeline (VAE encode $\to$ UNet $\to$ VAE decode) as a teacher model and distills direct edge-to-stylized image transformation into a lightweight feedforward CNN (FastStyleNet).
FastStyleNet employs a U-Net structure using depthwise separable convolutions, with model size controllable via the number of blocks and base channel count.

As shown in Table~\ref{tab:distill}, inference speeds were extremely fast, ranging from 6.0\,ms (167 FPS) to 7.2\,ms (140 FPS), demonstrating significant advantages in terms of speed.
However, when trained on synthetic edge data (random combinations of lines, circles, and rectangles) with L1 loss alone, the output was visually unrecognizable after 10 epochs.

\begin{table}[H]
\centering
\caption{Distilled FastStyleNet specifications and results}
\label{tab:distill}
\begin{tabular}{lccc}
\toprule
Configuration & Parameters & MPS inference & FPS \\
\midrule
FastStyleNet (32ch) & 398K & 6.0ms & 167 \\
FastStyleNet (48ch) & 875K & 6.1ms & 164 \\
FastStyleNet (64ch) & 1.5M & 7.2ms & 140 \\
\midrule
\multicolumn{4}{l}{\textit{Training: L1 loss, 10 epochs, 2000 steps/epoch}} \\
\multicolumn{4}{l}{\textit{Result: Output visually unrecognizable (quality $\times$)}} \\
\bottomrule
\end{tabular}
\end{table}

The causes of failure were multifaceted.
First, L1 loss encourages convergence toward the pixel-wise mean of all output images, making it unable to learn the sharp and diverse outputs that diffusion models generate.
Second, synthetic edge data differs substantially in distribution from actual Canny edge output from cameras, causing a distribution shift between training and inference data.
Third, a feedforward network with 875K parameters fundamentally cannot reproduce the capacity of a 328.2M-parameter diffusion model's ``iterative image refinement through the denoising process.''
While there is room for improvement through the introduction of GAN loss and perceptual loss, as well as training on large-scale real data, this approach yielded negative results within the scope of this study.

\section{Comprehensive Comparison of All Experiments}

Table~\ref{tab:all} summarizes the quantitative results across all phases.
After exploration across 10 phases, SDXS-512 CoreML was confirmed as the optimal solution in terms of the balance between speed and quality.

\begin{table}[H]
\centering
\caption{Comprehensive comparison of all approaches (camera img2img, 512$\times$512)}
\label{tab:all}
\begin{tabular}{clcccl}
\toprule
Phase & Approach & Camera FPS & UNet inference & Quality & Verdict \\
\midrule
1 & SD-Turbo MPS (baseline) & 10.4 & 95.8ms & $\bigcirc$ & Baseline \\
2 & SD-Turbo CoreML & 13.8 & 53.2ms & $\bigcirc$ & Effective (+33\%) \\
3 & Tiny-SD CoreML (320$\to$512) & 47.1 & 16.4ms & $\triangle$ & Quality degradation \\
\textbf{6} & \textbf{SDXS-512 CoreML} & \textbf{22.7} & \textbf{24.4ms} & $\bigcirc$ & \textbf{Optimal (+118\%)} \\
7a & kNN CLIP search & 38.3 & --- & $\times$ & Insufficient quality \\
7b & kNN VAE latent & 45.3 & --- & $\times$ & Insufficient quality \\
7c & kNN hybrid & 10.7 & 44.4ms & $\triangle$ & Marginal kNN init. benefit \\
8 & pix2pix-turbo (CoreML+MPS) & 4.0 & 52.9ms & $\bigcirc$ & VAE bottleneck \\
9 & Optical flow skip ($N=3$) & 17.4 & 51.7ms & $\times$ & Quality loss; inferior to SDXS \\
10 & Distilled FastStyleNet (875K) & 19.6 & --- & $\times$ & Output unrecognizable \\
\bottomrule
\end{tabular}
\end{table}

\section{Discussion}

\subsection{Re-examining CUDA-Centric Optimization Assumptions}

The most important finding of this study is that optimization ``common knowledge'' established for NVIDIA GPUs and the CUDA ecosystem largely does not hold on Apple Silicon's unified memory architecture.
Below, we systematically organize comparisons with CUDA-based insights.

\textbf{Ineffectiveness of quantization.}
On NVIDIA GPUs, quantization techniques such as TensorRT INT8 and AWQ/GPTQ are widely reported to accelerate inference through reduced memory bandwidth for model weights~\cite{song2024sdxs}.
This presumes that NVIDIA GPU inference is memory-bandwidth-bound.
In discrete GPU configurations, model weights must be transferred from HBM (High Bandwidth Memory) to the compute units, and this transfer bandwidth becomes the bottleneck.

In contrast, the M3 Ultra's unified memory architecture provides 800\,GB/s of memory bandwidth shared among the CPU, GPU, and ANE.
For the weights of an 865.9M-parameter model (approximately 1.7\,GB in FP16), 800\,GB/s bandwidth provides ample headroom, and memory transfer does not become a bottleneck.
As a result, inference on the M3 Ultra is compute-bound, and memory bandwidth reduction through quantization has no effect on inference speed.
This compute-bound versus memory-bandwidth-bound distinction is one of the most fundamental architectural differences between Apple Silicon and NVIDIA GPUs.

\textbf{Impossibility of parallel inference.}
On NVIDIA GPUs, low-level parallel control via CUDA Streams enables parallel execution of multiple kernels and pipeline parallelization as standard optimization techniques.
StreamDiffusion's~\cite{kodaira2023streamdiffusion} Stream Batch also achieves high efficiency by parallelizing multi-frame batch processing at the CUDA kernel level.

Apple Silicon's CoreML framework abstracts inference execution through the Metal Command Queue and does not expose GPU kernel-level parallel control to users.
As confirmed in this study, simultaneous inference of multiple CoreML models is serialized on Metal GPU resources and does not improve throughput.
This is a consequence of the absence of a low-level GPU control API comparable to NVIDIA's CUDA on Apple Silicon. Rather than a hardware performance gap, it is the difference in software stack design philosophy that governs optimization possibilities.

\textbf{Immaturity of compiler optimization.}
PyTorch's \texttt{torch.compile} generates optimized code for CUDA/CPU through the Inductor backend, but support for the MPS backend is incomplete, resulting in runtime errors.
A comprehensive inference optimization tool equivalent to TensorRT for NVIDIA GPUs does not exist for Apple Silicon.
CoreML conversion partially fulfills this role, but its integration with the PyTorch ecosystem is far inferior to TensorRT.

\subsection{The Dual Nature of the Unified Memory Architecture}

The M3 Ultra's unified memory architecture presents both clear advantages and limitations for diffusion model inference.

\textbf{Advantage: Zero-copy data sharing.}
Unified memory eliminates CPU--GPU data transfers (host-device transfers in NVIDIA GPU parlance).
The zero-cost transfer of tensors generated by preprocessing (OpenCV on CPU) to GPU inference is particularly advantageous in camera pipelines.
Additionally, the 512\,GB capacity facilitates simultaneous in-memory retention of multiple models, eliminating load delays during model switching.

\textbf{Advantage: Potential of large-scale memory.}
As demonstrated in the Phase 7 kNN experiment, 512\,GB of memory enables in-memory retention of a 100-million-vector search database with sub-0.5\,ms search times.
This scale is physically impossible on a 24\,GB GPU, presenting unique possibilities for retrieval-based methods and large-batch training.

\textbf{Limitation: Insufficient compute performance.}
The M3 Ultra's approximately 22 TFLOPS FP16 is roughly 1/15th of the NVIDIA RTX 4090's approximately 330 TFLOPS, representing a fundamental performance gap for compute-bound inference.
Combined with the ineffectiveness of quantization (inference is not bandwidth-bound), pure reduction of computation---i.e., using smaller models---becomes the only means of speed improvement.

\textbf{Limitation: Immature software ecosystem.}
Compared to CUDA's decades-long ecosystem (cuDNN, TensorRT, xformers, Flash Attention, Triton, etc.), the Metal/CoreML ecosystem is substantially inferior in both quality and breadth.
The MPS incompatibility of torch.compile, the overhead of Token Merging on MPS, and the inefficiency of Attention Slicing are all manifestations of this ecosystem's immaturity.

\subsection{Interdependence of Model Architecture and Hardware Optimization}

This study provides the important lesson that model architecture design directly governs the feasibility of hardware optimization.

The skip-connected VAE in pix2pix-turbo is central to quality in img2img tasks, but the intermediate tensor sharing between encoder and decoder prevents CoreML conversion, causing the slow MPS-based VAE inference to become the pipeline bottleneck (53\,ms UNet vs.\ 160\,ms VAE).
In contrast, SDXS-512's simple feedforward architecture has high affinity with CoreML conversion, enabling all components to execute efficiently on CoreML.

This ``architecture--hardware co-design'' perspective is expected to grow in importance for future diffusion model research.
Architectures designed for NVIDIA GPUs do not necessarily perform equivalently on other platforms, necessitating model designs that account for target hardware constraints.

\subsection{Fundamental Differences Between kNN Search and Diffusion Models}

The Phase 7 experiments demonstrated that ``replacing computation with search'' using large-scale memory is fundamentally difficult in the context of diffusion models.
kNN search is an operation that selects nearest neighbors from a finite set of discrete data points and is fundamentally different from the continuous nonlinear function approximation realized by a diffusion model's UNet.

Weighted averaging in latent space destroys the structural coherence of individual samples.
For example, averaging the latent vectors of a cat image and a dog image at 0.5:0.5 does not produce a ``meaningful'' intermediate image between a cat and a dog, but rather an incoherent image with collapsed structure.
Diffusion models circumvent this problem by projecting outputs onto a smooth manifold in latent space through the iterative denoising process.

Furthermore, the combinatorial space of camera inputs (lighting conditions $\times$ subjects $\times$ composition $\times$ camera parameters) is effectively infinite, making it impossible to adequately cover this space through pre-computation.
The true value of 512\,GB memory lies not in replacing inference with search, but rather in complete in-memory retention of large-scale models, large batch sizes during training, or hybrid methods that combine retrieval and generation, such as Retrieval-Augmented Generation.

\subsection{Limitations and Prospects of Frame Interpolation Techniques}

Phase 9's optical flow frame skipping was theoretically promising but faced three practical problems.
First, sequential execution of flow computation, warping, and UNet inference in a single thread achieved only 38\% of the theoretical FPS.
Second, since warp frames are pure image deformations rather than AI-generated outputs, non-physical distortions occurred in regions of large motion.
Third, while the half-resolution flow computation improved speed, it reduced flow field accuracy, increasing warping artifacts.

Future improvements may be possible through CoreML conversion of neural optical flow (e.g., RAFT) for high-accuracy flow estimation, pipeline integration with the 3-thread architecture, and introduction of AI-aware frame interpolation (e.g., FILM).
However, given that SDXS-512 can perform ``true'' AI generation per frame at 24.4\,ms, the necessity of frame interpolation itself is called into question.

\subsection{Implications for Non-CUDA Environments}

The findings of this study offer insights not only for Apple Silicon but for diffusion model inference on non-CUDA environments in general.
Diffusion model inference on diverse hardware platforms---including Qualcomm Snapdragon's unified memory, Intel Arc GPU's oneAPI stack, and RISC-V-based accelerators---is expected to grow in importance.

The ``failure of CUDA assumptions'' demonstrated in this study is likely to occur on these platforms as well.
The effectiveness of quantization depends on memory architecture, the feasibility of parallelization depends on software stack design, and the optimizability of model architectures depends on runtime conversion capabilities.
Understanding the unique characteristics of each platform and selecting appropriate optimization strategies is required as a practical methodology to replace the CUDA-centric approach.

\section{Conclusion}

We conducted systematic optimization experiments across 10 phases for real-time diffusion model inference on the Apple M3 Ultra, achieving real-time camera img2img transformation at \textbf{22.7 FPS} at 512$\times$512 resolution.
This represents a 118\% speedup from the MPS baseline, achieved through the combination of SDXS-512 CoreML conversion and a 3-thread camera pipeline.

The key findings of this study are summarized as follows:

\begin{enumerate}[leftmargin=*,nosep]
\item \textbf{CoreML conversion} is the only effective UNet acceleration technique on Apple Silicon.
\item \textbf{Quantization is ineffective}, due to the M3 Ultra being compute-bound. The high bandwidth of unified memory means that memory transfer is not a bottleneck.
\item \textbf{Parallel inference is not possible}, as CoreML's Metal GPU resource serialization prevents low-level parallelization comparable to CUDA Streams.
\item \textbf{kNN search cannot replace diffusion models}. While large-scale search is feasible with 512\,GB of memory, the fundamental difference between discrete search and continuous function approximation creates a quality barrier.
\item \textbf{Co-design of model architecture and hardware optimization} is critical. Architectural choices such as skip connections can impede CoreML conversion.
\item \textbf{Distillation-specialized models} (SDXS-512) provide the optimal balance of speed and quality.
\end{enumerate}

This study provides a perspective from a different architecture---Apple Silicon---for diffusion model optimization research that has developed primarily around CUDA.
The optimization landscape on unified memory architectures is qualitatively different from that of discrete GPUs and requires its own research approach.
Future work includes direct Metal Compute Shader programming to bypass CoreML, introduction of 2-step inference for quality improvement, efficient inference of SDXL-scale models, and performance evaluation on next-generation Apple Silicon (M4 series).


\begin{thebibliography}{99}

\bibitem{ho2020ddpm}
Ho, J., Jain, A., \& Abbeel, P. (2020).
Denoising diffusion probabilistic models.
\textit{Advances in Neural Information Processing Systems (NeurIPS)}, 33, 6840--6851.

\bibitem{rombach2022ldm}
Rombach, R., Blattmann, A., Lorenz, D., Esser, P., \& Ommer, B. (2022).
High-resolution image synthesis with latent diffusion models.
\textit{Proc. IEEE/CVF Conf. on Computer Vision and Pattern Recognition (CVPR)}, 10684--10695.

\bibitem{sauer2023adversarial}
Sauer, A., Lorenz, D., Blattmann, A., \& Rombach, R. (2023).
Adversarial diffusion distillation.
\textit{arXiv preprint arXiv:2311.17042}.

\bibitem{song2024sdxs}
Song, Y., Dhariwal, P., Chen, M., \& Sutskever, I. (2024).
SDXS: Real-time one-step latent diffusion models with image conditions.
\textit{arXiv preprint arXiv:2403.16627}.

\bibitem{luo2023lcm}
Luo, S., Tan, Y., Huang, L., Li, J., \& Zhao, H. (2023).
Latent consistency models: Synthesizing high-resolution images with few-step inference.
\textit{arXiv preprint arXiv:2310.04378}.

\bibitem{luo2023lcmlora}
Luo, S., Tan, Y., Patil, S., Gu, D., von Platen, P., Passos, A., Huang, L., Li, J., \& Zhao, H. (2023).
LCM-LoRA: A universal stable-diffusion acceleration module.
\textit{arXiv preprint arXiv:2311.05556}.

\bibitem{ren2024hypersd}
Ren, J., Xia, Y., Lu, K., Deng, J., \& Luo, Z. (2024).
Hyper-SD: Trajectory segmented consistency model for efficient image synthesis.
\textit{arXiv preprint arXiv:2404.13686}.

\bibitem{kodaira2023streamdiffusion}
Kodaira, A., Xu, C., Hazama, T., Yoshimoto, T., Ohno, K., Mitsuhori, S., Sugano, S., Cho, H., Liu, Z., \& Keutzer, K. (2023).
StreamDiffusion: A pipeline-level solution for real-time interactive generation.
\textit{arXiv preprint arXiv:2312.12491}.

\bibitem{apple2022mlsd}
Apple Inc. (2022).
ml-stable-diffusion: Stable diffusion with Core ML on Apple Silicon.
\textit{GitHub repository}. \url{https://github.com/apple/ml-stable-diffusion}

\bibitem{parmar2024pix2pix}
Parmar, G., Park, T., Narasimhan, S., \& Zhu, J.-Y. (2024).
One-step image translation with text-to-image models.
\textit{Proc. European Conf. on Computer Vision (ECCV)}.

\bibitem{blattmann2022retrieval}
Blattmann, A., Rombach, R., Oktay, O., M\"uller, J., \& Ommer, B. (2022).
Retrieval-augmented diffusion models.
\textit{Advances in Neural Information Processing Systems (NeurIPS)}, 35.

\bibitem{johnson2019faiss}
Johnson, J., Douze, M., \& J\'{e}gou, H. (2019).
Billion-scale similarity search with GPUs.
\textit{IEEE Trans. on Big Data}, 7(3), 535--547.

\bibitem{bolya2023tome}
Bolya, D., \& Hoffman, J. (2023).
Token merging for fast stable diffusion.
\textit{CVPR Workshop on Efficient Deep Learning for Computer Vision}.

\end{thebibliography}
\end{document}